\documentclass[twocolumn,runningheads]{llncs} 

\usepackage{geometry}
\usepackage[T1]{fontenc}

\usepackage{tabularx}
\usepackage{booktabs}
\usepackage{subcaption}
\usepackage{graphicx}
\usepackage{hyperref}
\usepackage{amssymb}

\usepackage{color}

\begin{document}
\title{A Vision Based System for Guided and Collaborative Reconstruction of Fragmented Documents}

\titlerunning{Vision-Based Document Reconstruction}
%
\author{Oliver Krumpek \and
Diana Leo}
\authorrunning{O. Krumpek, D. Leo}
%
\institute{Fraunhofer IPK, dept. Machine Vision, Pascalstr. 8-9, 10587 Berlin, Germany}
\maketitle              
\begin{abstract}
This paper presents the development and evaluation of a collaborative system for real-time reconstruction of fragmented paper documents in the context of cultural heritage preservation. 
The developed system includes a collaborative robot, or cobot, that can fully manage the positioning of paper fragments using a specially designed vacuum-based suction attachment. This attachment enables gentle and precise positioning, ensuring the preservation of fragile materials. With this device, we are able to achieve a positioning repeatability of 0.57mm for fragments of 8cm². The system offers users the flexibility to choose between manual positioning, with visual guidance, or fully automated positioning performed by the cobot. To further improve the reconstruction process, AI methods for image interpretation, specifically for segmentation and positioning tasks, were applied and evaluated for their applicability to template-based reconstruction of damaged paper fragments. Our investigation provides critical insights into the performance of different local feature matching methods under different document types, taking into account rotation, scale robustness, and the degree of damage to the fragments. With a focus on the reconstruction of damaged and optically altered archival material, SE2-LoFTR, a detector-free local feature matching method, was chosen as the preferred method for the system due to its robust performance in our experiments.

\keywords{Automatic Reconstruction \and Assistance System \and Fragmented Paper \and Feature Matching \and Human Machine Interaction \and Collaborative Systems}
\end{abstract}

\section{Motivation}

The Fraunhofer Institute for Production Systems and Design Technology (IPK) is part of the Research Alliance for Cultural Heritage (FalKe), which comprises the Fraunhofer-Gesellschaft, the Leibniz Association and the Prussian Cultural Heritage Foundation \cite{FalKe.23}. We claim that the use of automation technology, including collaborative robots (cobots), has the potential to contribute to the feasibility of reconstruction projects. This is particularly true for projects with high repeatability of the handling tasks. Using advanced sensors and vision systems, cobots can carefully navigate and manipulate delicate materials, taking over the task of positioning fragments of different types and sizes. The integration of computer vision and artificial intelligence for automatic image analysis further enhances the capabilities for accurate reconstruction. The use of automation technology streamlines and accelerates the reconstruction process, resulting in improved efficiency.

\begin{figure}[htb]
\centering
\includegraphics[width=0.48\textwidth]{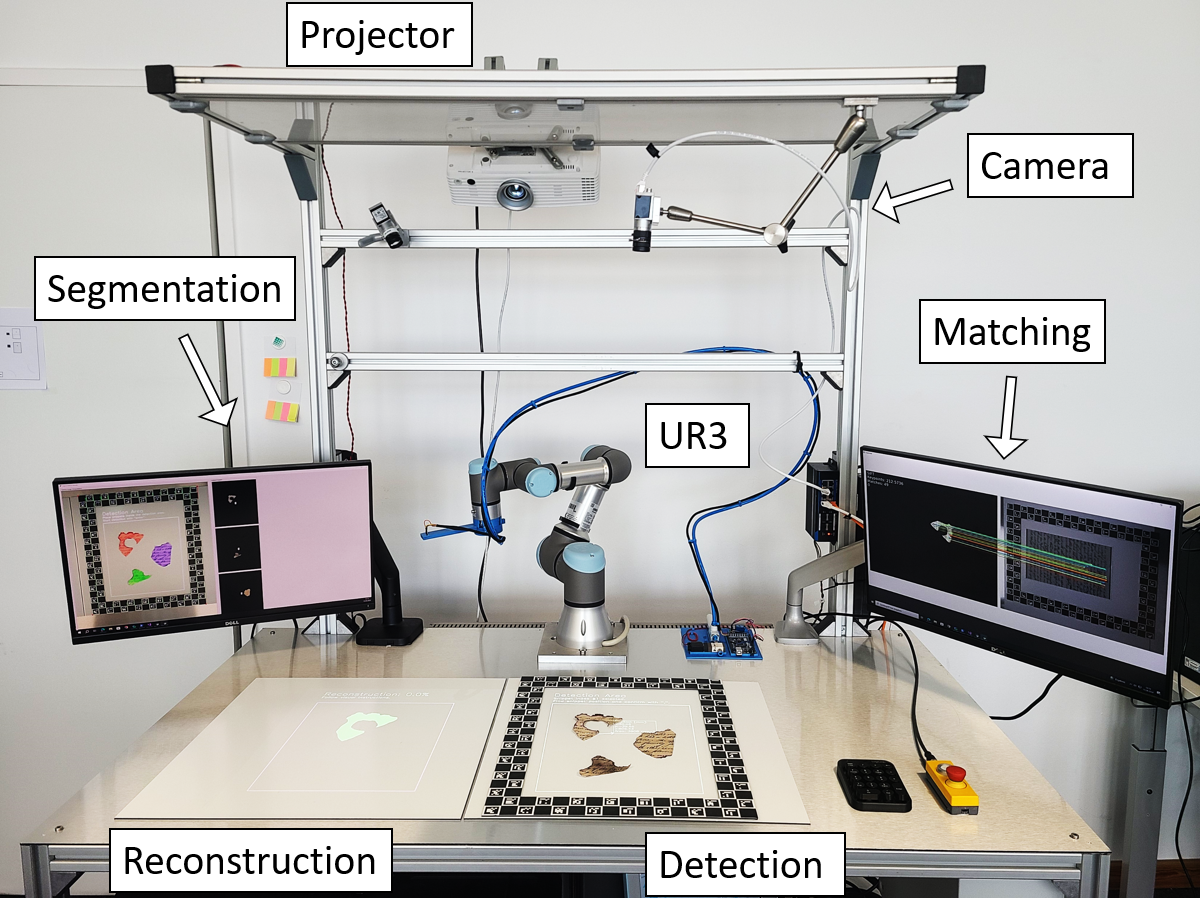}
\caption{Annotated image of the overall hardware setup, illustrating the key components. The system includes an industrial camera that digitises the fragments, and a projector that displays the user interface, providing real-time feedback during the reconstruction process. The user can either manually place the fragment in the found target position or delegate this task to the centrally positioned cobot, which handles the precise placement upon request.}
\label{fig:hardware_setup}
\end{figure}

Assistive systems that combine human expertise with machine capabilities allow experts to focus on more complex aspects of the reconstruction process such as process control, correction and validation. This human-machine synergy improves the overall reconstruction process and contributes to the preservation of cultural artefacts. Furthermore, the application of collaborative robotics and automation technology extends beyond the reconstruction of paper documents. The principles and methodologies developed in this context can be adapted  to the restoration of works of art, the reconstruction of archaeological artefacts and the preservation of historical manuscripts.

The system introduced in this study was initially inspired by a case study \cite{FalKe.23} exploring the potential of automation. In its current form, the system employs a template-based puzzle-solving method specifically designed for the reconstruction of documents, such as those typically found in archives and libraries. However, this method is designed as an interchangeable module, allowing for future replacement with more advanced, fully automatic algorithms that do not require reference templates. The primary objective of this research is to develop a robust reconstruction approach tailored to the specific use case where a reference is available. 

In conclusion, the integration of collaborative robots and automation technology in the reconstruction of cultural artefacts offers opportunities for increased efficiency and accuracy. By leveraging advanced technologies and combining human expertise with machine capabilities, assistive systems contribute to the preservation and appreciation of our shared cultural heritage.

\section{Contributions}

\begin{figure}
\centering
\includegraphics[width=0.48\textwidth]{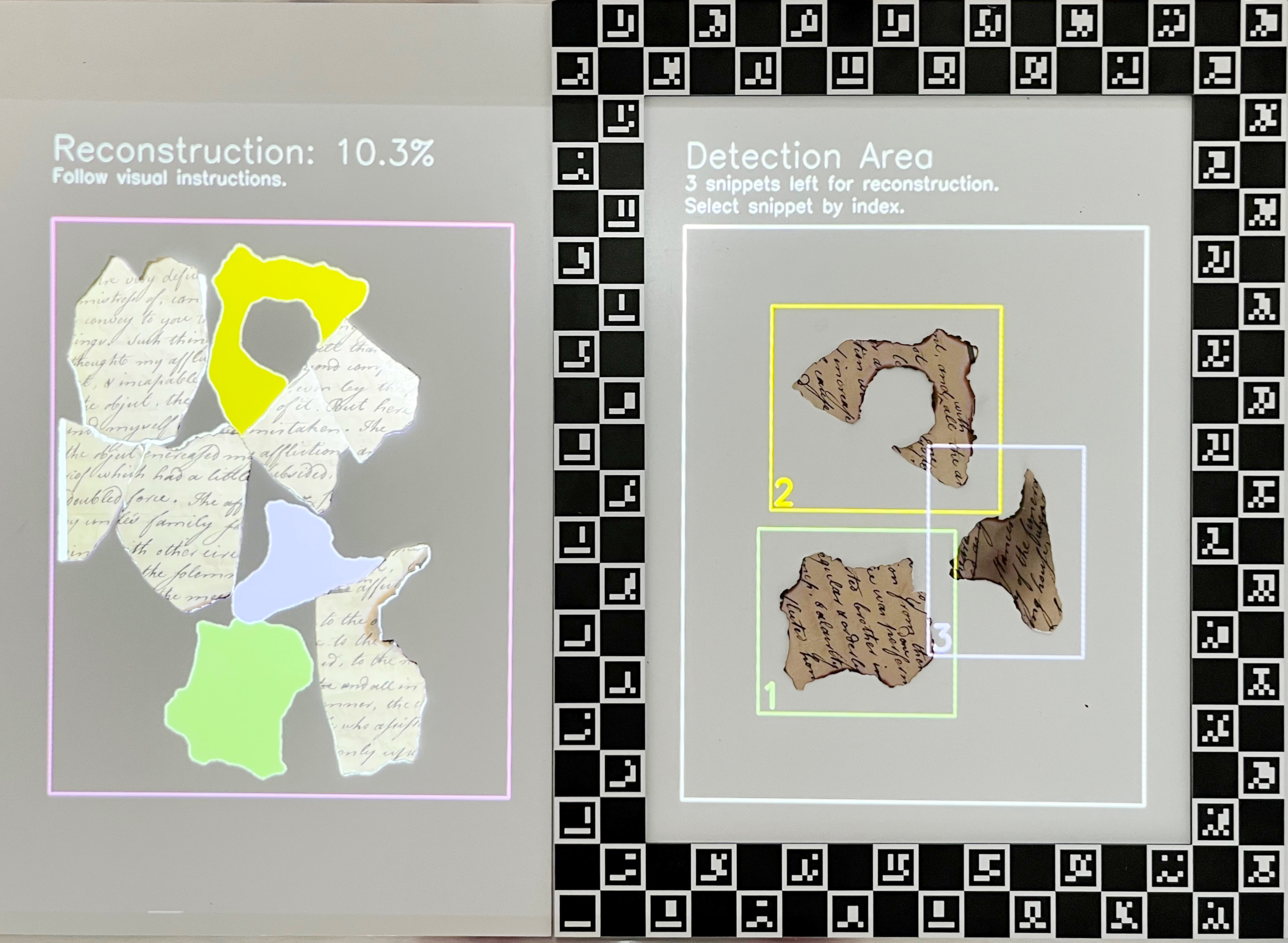}
\caption{
Photo of the projected user interface for a guided reconstruction process. 
Users place snippets in the so-called detection area where paper fragments are digitised using an industrial camera and processed using AI methods for background subtraction and feature matching. The detected fragments are framed in highlighted squares (right). The virtually assembled document is visualized in the so-called reconstruction area. By selecting the identified fragments by their index, the system highlights their orientation and position, providing visual guidance for the manual reconstruction (left).}
\label{fig:UI}
\end{figure}

\begin{itemize}
    \item \textbf{Contribution 1} : We have developed a partially automated system for guided reconstruction of damaged or corrupted cultural artefacts in the form of flat fragments, such as paper documents or mosaic pieces. The proposed system combines a collaborative robot, data-driven vision systems and a projector-based interface to facilitate human-machine collaboration in the reconstruction task (see Fig. \ref{fig:UI}). Our functional and reproducible system design enables widespread adoption and use.\\

    \item \textbf{Contribution 2} : We have developed a material-protective, vacuum-based suction attachment equipped with sensors to safely handle paper fragments with minimal physical impact. The device can be either used handheld by the restorers or mounted at the Tool Center Point (TCP) of collaborative robots. We are making this attachment publicly available for widespread use, promoting collaboration and innovation in the preservation of document fragments.\\
    
    \item \textbf{Contribution 3} : Our third contribution is the evaluation of local feature matching methods for template-based document reconstruction in assistance systems. Specifically, we assessed three local feature matching methods (SIFT, SuperPoint+SuperGlue, and SE2-LoFTR) for their effectiveness in reconstructing damaged paper fragments. We evaluated their suitability based on robustness to rotation, scaling and damage intensity. Our results highlight both the strengths and limitations of these methods, providing valuable insights for selecting appropriate techniques in similar system setups.
\end{itemize}

\section{Related Work} \label{sec:relatedwork}

The concept of human-robot collaboration (HRC) has gained significant attention, particularly in the industrial sector. HRC focuses on creating cooperative partnerships between humans and robots, where they work together to accomplish tasks that leverage their respective strengths.\cite{Schmidbauer.2021} Depending on the specific task, this collaborative approach allows for increased productivity, improved safety, and enhanced efficiency in various industries and use cases. This is particularly true in areas such as reverse logistics and the circular economy, where the automated task of assembly and disassembly is becoming increasingly important. \cite{Daneshmand.2023}. In addition to or in combination with HRC-systems, data driven worker assistance systems (WAS) can help to decrease the workload within increasingly complex tasks by providing the right amount of information in the right form at the right time.\cite{Bornewasser.2018}.

In the field of cultural heritage preservation, robotic-assisted systems have long been used for digitisation and virtual reconstruction tasks\cite{Luxman.2022} \cite{HardyHelen2020Radi}. Furthermore, the potential usage of robotic systems for physical reconstruction has already been shown by the RePAIR project \cite{ComputerVisionNews-2021Octobe}. This initiative specifically targets the restoration of ancient artworks in Pompeii, leveraging shape, 3D information, and decorative elements to establish connections among fragmented pieces, and using soft robotics for delicate handling and manipulation tasks. Corresponding research also exists for the general use of robots for the repair of fractured objects based on methods of reinforcement learning \cite{Song.10.02.202312.02.2023}. Similar to our system setup, the authors of\cite{Ma.2023} also used template-based matching and a robot manipulator while dealing with solid puzzle pieces. Notably, new and promising developments in the field of flexible, shape-shifting grippers and manipulators based on principles such as electroadhesion, vacuum or pneumatic \cite{Zaidi.2021}, enable robotic systems to be used more extensively for tasks involving fragile and valuable objects.

More broadly, the scientific community has extensively explored automated solutions for fragment reassembly, focusing largely on virtual reconstruction without the use of robotic systems. This exploration encompasses not only theoretical exercises in computer vision and pattern recognition \cite{jig_puzz_robo,BUNKE19931797,goldberg.2002} but also practical challenges in reconstructing real-world 3D and 2D-items, like wall paintings \cite{wall_painting,arch_fragment,papaodysseus2008image} or shredded documents used in forensic investigations ~\cite{forensics,desmet.2009}. The complexity of those reassembly-tasks hinges on whether the original image is available. These are divided into template-based approaches where relationships between fragments are known \cite{Cai.12032016}, simplifying alignment and identification processes, and others where no original image exists, often addressed as Jigsaw-Puzzles. In this field, it has recently been shown that graph and diffusion models can be used to solve reconstruction tasks. In the context of the RePair project\cite{ComputerVisionNews-2021Octobe}, the authors of \cite{scarpellini2024diffassembleunifiedgraphdiffusionmodel} have just introduced DiffAssemble, which uses a unified model approach for 2D and 3D tasks, showing promising results on public datasets.

Li et al. \cite{banknotes} have demonstrated the effectiveness of local feature matching for reconstructing banknotes from numerous fragments, employing techniques that align and group fragments through agglomerative clustering.

\section{Methods and Materials}

In this section, we present the methods employed in our investigation to enable guided and automatic reconstruction of fragmented paper documents.

\subsection{Vision and Feature Matching Methods}

In order to identify the target position of the paper fragment in the reference template, our system uses proven and well-described algorithms for the extraction and matching of local features.

\begin{figure}
\centering
\includegraphics[width=0.48\textwidth]{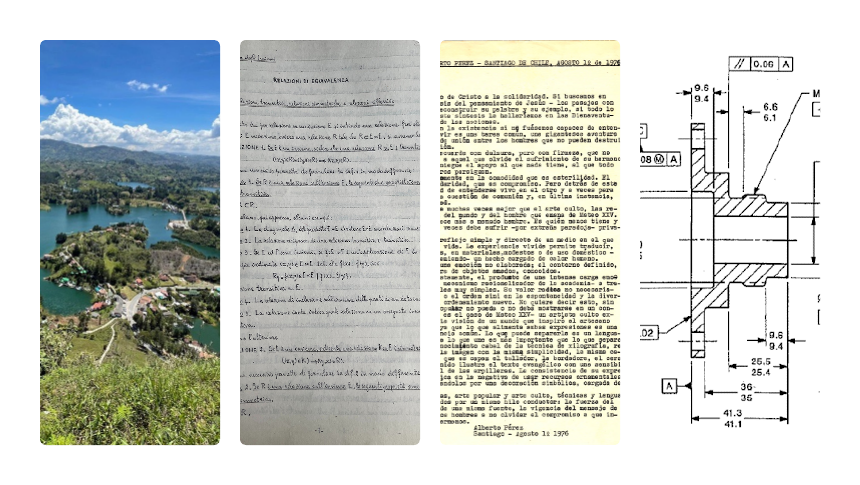}
\caption{Examples of document types used to evaluate the reconstruction system. From left to right: landscape image, handwritten document, typewritten document, blueprint.}
\label{fig:references}
\end{figure}

We implemented and evaluated three local feature matching methods: SIFT + Nearest Neighbor Matching (NN) (BF-Matcher or FLANN Matcher), SuperPoint+SuperGlue and SE2-LoFTR. The selection of these methods has been informed by research into traditional and state-of-the-art local feature matching methods. Given the requirements of our system, we focused on algorithms with strong rotation- and scale-invariant feature extraction capabilities. SIFT~\cite{sift} was chosen for its robustness in challenging scenarios involving illumination, scale, and rotation. SuperPoint~\cite{superpoint} was implemented for its advanced keypoint detection and description capabilities, especially when combined with SuperGlue~\cite{superglue}, known for its precision and reliability in feature correspondence. SE2-LoFTR~\cite{se2-loftr}, a detector-free method, was chosen for its inherent rotation invariance and streamlined workflow. Prior to the actual virtual positioning, we use the Neural Network based framework REMBG \cite{Daniel.Gatis2021} for automatic background removal and robust fragment segmentation.\\

\begin{figure}
\centering
\includegraphics[width=0.48\textwidth]{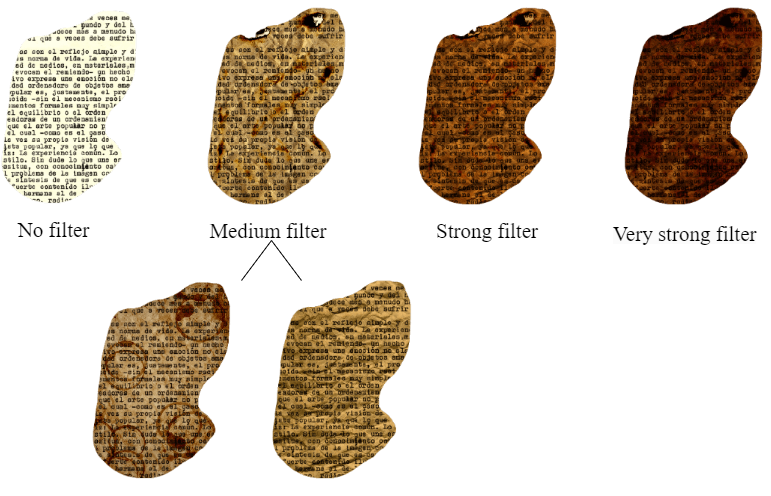}
\caption{Fragments with simulated damage created using filters that mimic varying levels of burns, stains, and crumpling. The medium damage level includes three variants, each simulating either burns, stains, or crumpling. These filters were applied to evaluate the system's performance in reconstructing documents under different damage conditions.}
\label{fig:filterlevel}
\end{figure}

In order to select the most appropriate and robust method for our specific application, we conducted a series of experiments to test the performance of the selected methods on fragments with increasing levels of damage. These experiments focused on the system's potential use for reconstructing damaged and contaminated fragments, such as those affected by natural or man-made disasters. To facilitate this, we created a dataset of digitally cut fragments with varying degrees of damage by adding filters that mimic burns, stains, or crumple (see Fig. \ref{fig:filterlevel}).

Four distinct types of documents were selected as reference images, as shown in Fig. \ref{fig:references}: handwritten, typewritten, blueprint and outdoor landscape. The choice of these document types was strategic: typewritten and handwritten documents are characterized by their highly repetitive patterns. In contrast, the blueprint documents exhibit sparser features, primarily composed of distinct geometric shapes and lines, presenting a different set of challenges for feature matching methods due to their less textured nature. The outdoor landscape, representing the standard imagery of natural scenes, is classified as the easy document type because it offers a broader variety of distinctive features like varying textures and natural elements. These documents were selected because the system is intended to operate in the field of cultural heritage preservation, specifically targeting archives and libraries.

\subsection{Setup and Hardware}

\begin{figure}
\centering
    \includegraphics[width=0.38\textwidth]{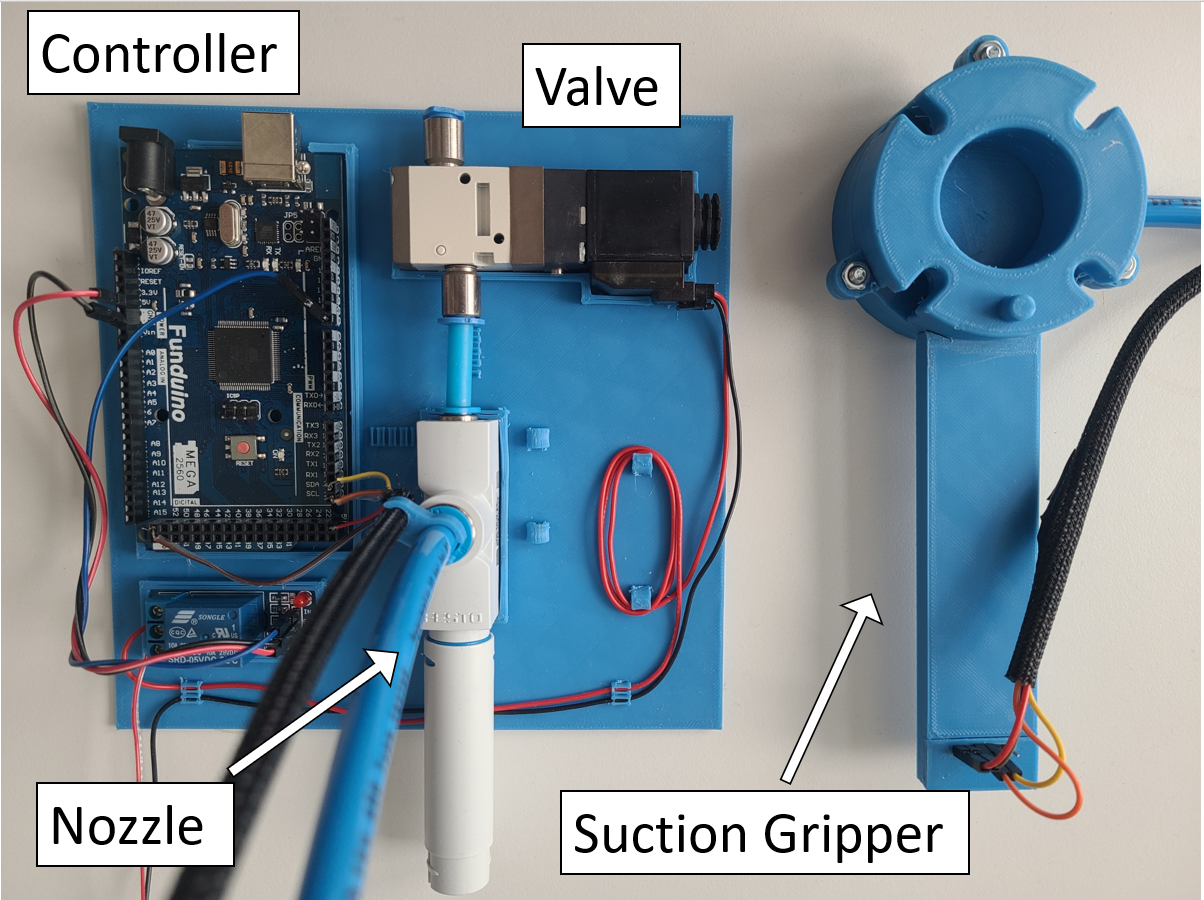}
  \caption{Overview of mechanical and electrical components of the suction gripper. The control unit and the valve are rigidly attached to a printed carrier board and connected to the suction gripper using flexible tubing and isolated wires.}
    \label{fig:electronics}
\end{figure}

\begin{figure}
\centering
\begin{subfigure}{0.23\textwidth}
    \includegraphics[width=\textwidth]{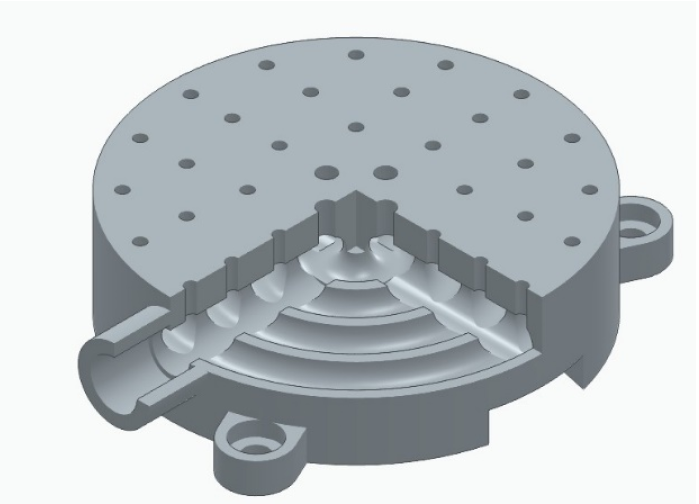}
    \caption{Rendering of the suction attachment with insights to the inner structure}
\end{subfigure}  
\hspace{0.01\textwidth}
\vspace{0.02\textwidth}
  \begin{subfigure}{0.23\textwidth}
    \includegraphics[width=\textwidth]{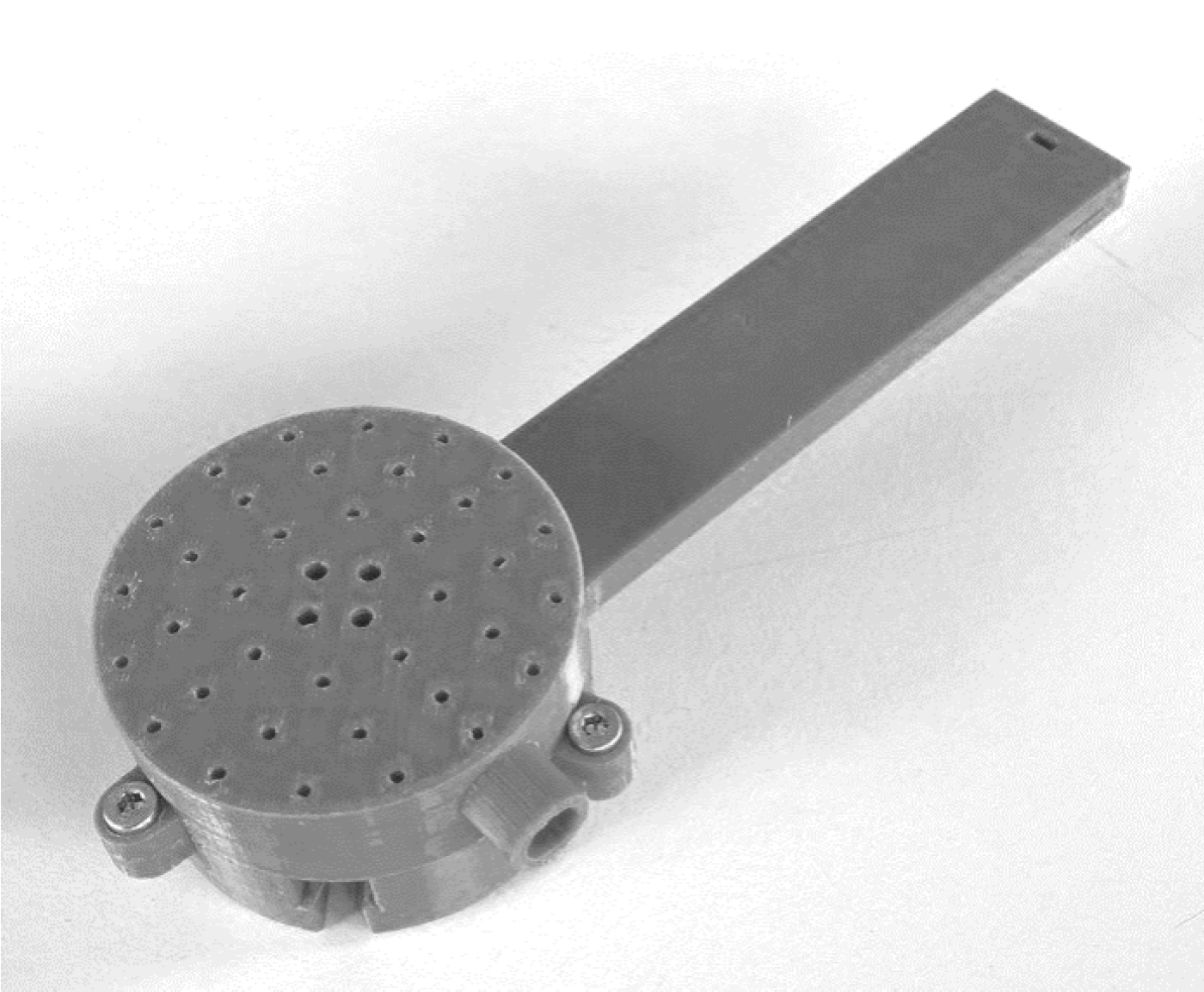}
    \caption{Printed version of the suction attachment for the use on UR5 endeffector}
  \end{subfigure} 
  \caption{Design of the suction gripper}
    \label{fig:Design of the suction gripper}
  \end{figure}

For automatic handling of fragments with the cobot, a gripping mechanism was necessary. Therefore, a vacuum gripper was developed to gently pick up the paper fragments without causing additional damage. This device was designed to handle fragments of varying sizes (see Fig. \ref{fig:Design of the suction gripper}). Initially, we experimented with a sponge-based approach, but this was eventually replaced by a 3D-printed structure featuring a balanced distribution of air channels, determined through experimentation (see Fig. \ref{fig:Design of the suction gripper}). The gripper is equipped with a proximity sensor (VL53L0X Time-of-Flight) that allows the airflow to be automatically activated just before contact with the fragment, ensuring effectiveness for both pick-up and deposit processes. The required vacuum is generated and controlled using a Festo VN-14-H vacuum generator and a 3V210-08 24VDC valve (see Fig. \ref{fig:electronics}).

\subsection{System Calibration and Position Measurement}

\begin{figure}
\centering
\includegraphics[width=0.38\textwidth]{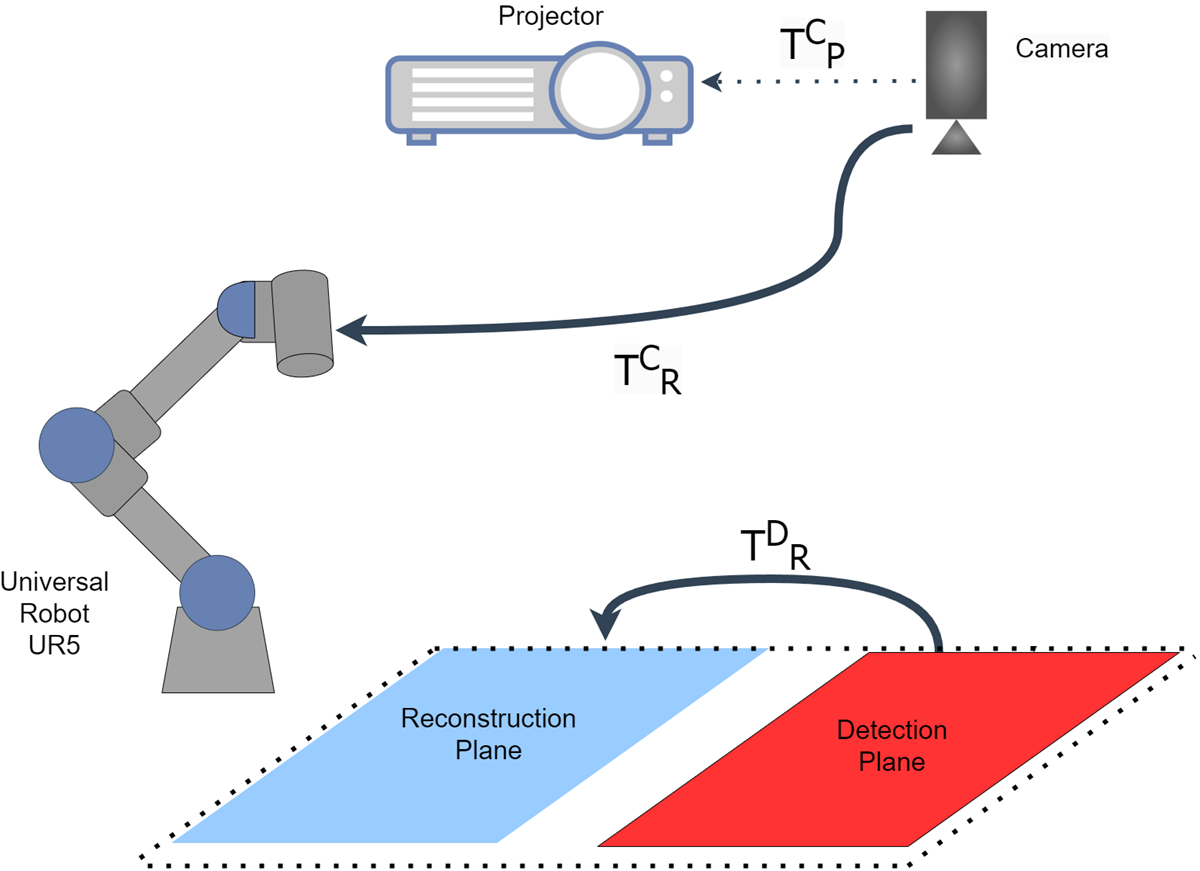
}
\caption{System diagram illustrating the key geometric transformations, including the camera-to-robot transformation $T^C_R$, the camera-to-projector homography $T^C_P$ and the transformation between the initial fragment position in the detection-area and its target position in the reconstruction-area $T^D_R$.}
\label{fig:transformations}
\end{figure}

In order to determine the geometric relationships of the components used in the system (see Fig.\ref{fig:transformations}), several steps were first taken to calibrate the system, using essentially straightforward methods. 

The relevant geometrical relations for the coordinate transformation performed during the evaluation are given by $T^C_R$, the constant transformation between camera and robot and $T^C_P$, the constant transformation between camera and projector. Both transformations can be carried out in a one-off calibration prior to the live evaluation. To calculate $T^C_R$, a system of the form $AX=YB$, also addressed as a hand-eye-calibration \cite{daniilidis1999hand} was set up and solved.

To enable the transformation between camera and projector $T^C_P$ we first assumed all fragments to be planar as the problem could then be reduced to the estimation of a homography between two planes and thus determined by the following expression: $(\lambda x',\lambda y',\lambda )^T = H(x,y,1)^T$ 
Here, $H$ describes the 2D to 2D projective transformation between the 2D plane in the 3D world and the 2D image plane. To solve for $H$, we find at least four corresponding points between the camera image and the projected image. This is achieved by projecting a calibration pattern onto the detection surface and detecting it in the camera image.\\

For manual assisted reconstruction, it was first necessary to express a description of significant features such as edges, centroids and surface information in terms of pixel values. These values could be processed for the visual interface based on the projector-to-camera calibration described above. 
However, in order to implement automatic positioning, it was necessary to derive positions in Euclidean space in addition to the pixel values, on the basis of which the cobot`s path planning could be performed. To keep the hardware setup simple and reproducible, we avoided using an additional camera or a 3D sensor. Instead, we used a ray tracing approach to determine the 3D position of the fragments' centre of gravity. More precisely, given the intrinsic camera parameters\cite{Tsai}, the desired 3D coordinate is given by intersecting the ray from the camera's optical centre through the pixel position $(u,v)$ with the plane defined by at least three 3D points (we use three corner points) of the ChAruco frame placed around the detection surface, all expressed in the Euclidean camera coordinate system (see Fig. \ref{fig:hardware_setup}). 

\section{Results}
\begin{figure}
\centering
\includegraphics[width=0.48\textwidth]{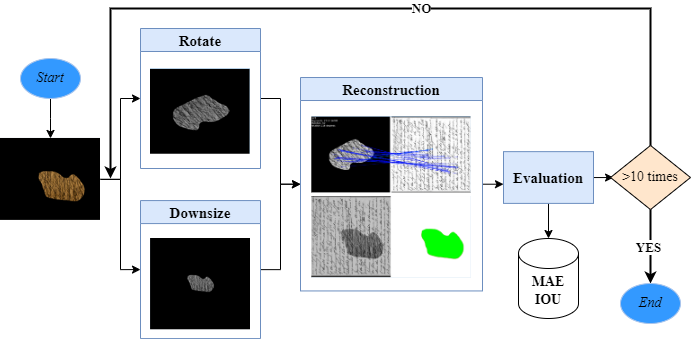}
\caption{Schematic representation of the experimental pipeline detailing the steps involved in the evaluation of the reconstruction of rotated (Experiment 1) or scaled (Experiment 2) document fragments. Ten rotations and ten levels of down-scaling are applied to the fragment to simulate environmental conditions, e.g. camera offsets.}
\label{fig:pipeline}
\end{figure}

Our comprehensive evaluation across two robustness experiments— Rotation and Scale—showcases the varied performance of the three selected local feature matching methods under multiple damage conditions (see Fig. \ref{fig:filterlevel}). The evaluated local feature matching methods are: traditional SIFT+NN, SuperPoint+NN, SuperPoint+SuperGlue, and detector-free SE2-LoFTR. Each damaged fragment from the created dataset goes through the pipeline illustrated in Fig. \ref{fig:pipeline}. The results are categorized into three performance levels based on the Intersection Over Union (IOU) and Mean Absolute Error (MAE): "Good (MAE = [0,8]px)", "Moderate (MAE = [9,22]px)", and "Failed (MAE$>$22px)". 
The tables \ref{tab:failure_rate_sift}, \ref{tab:failure_rate_se2} and \ref{tab:failure_rate_sg} present the reconstruction failure rates SIFT, SE2-LoFTR and SuperPoint+SuperGlue respectively, across the different document types, experiments, and filter levels (medium, strong, and very strong).

\subsection{Rotation Robustness Experiment}
Each snippet, already subjected to four increasing damage levels (no damage, medium damage, hard damage, very hard damage) (see Fig. \ref{fig:filterlevel}), was rotated through ten angles ranging from 0 to 324 degrees in increments of 36 degrees.
SE2-LoFTR and SIFT demonstrated the most robust performance in rotation invariance across various document types and levels of damage, showcasing their inherent rotation invariance. Notably, traditional SIFT often outperformed SE2-LoFTR, particularly with less damaged documents, showcasing its enduring effectiveness despite advancements in deep learning methods like SE2-LoFTR. SuperPoint+NN and SuperPoint+SuperGlue showed more limited robustness, excelling under certain conditions but struggling in others. The handwritten document was the most challenging to reconstruct, followed by the technical drawing. The landscape document was the easiest, with only a 2\% failure rate. The handwritten document posed significant challenges for the SuperPoint detector, regardless of the matcher used. Technical drawings resulted particularly difficult for SuperPoint+NN, and the most difficult to reconstruct for SE2-LoFTR.

\begin{table}
\centering
\caption[Failure rate by document type for SIFT]{Failure rate by document type for SIFT.}
\label{tab:failure_rate_sift}
{ 
\begin{tabularx}{0.49\textwidth}{X||c|c|c|c|c|c}
\toprule
Document & \multicolumn{2}{c|}{Medium} & \multicolumn{2}{c|}{Strong}& \multicolumn{2}{c}{Very Strong} \\
\cmidrule{2-7}
& Rotation & Scale & Rotation & Scale & Rotation & Scale \\
\midrule
Handwritten & 0\% & 40\% & 0\% & 60\% & 100\% & 100\%\\ 
Typewritten & 0\% & 30\% & 0\% & 60\% & 0\% & 70\%\\ 
Blueprint& 0\% & 50\% & 0\% & 70\% & 0\% & 70\%\\ 
Landscape & 0\% & 20\% & 0\% & 30\% & 0\% & 70\%\\ 
\bottomrule
\end{tabularx}}
\end{table}

\begin{table}
\centering
\caption[Failure rate by document type for SE2-LoFTR]{Failure rate by document type for SE2-LoFTR.}
\label{tab:failure_rate_se2}
{ \begin{tabularx}{0.49\textwidth}{X||c|c|c|c|c|c}
\toprule
Document & \multicolumn{2}{c|}{Medium} & \multicolumn{2}{c|}{Strong}& \multicolumn{2}{c}{Very Strong} \\
\cmidrule{2-7}
& Rotation & Scale & Rotation & Scale & Rotation & Scale \\
\midrule
Handwritten & 0\% & 30\% & 20\% & 50\%  & 40\% & 60\% \\ 
Typewritten & 0\% & 50\% & 10\% & 80\% & 10\% & 80\%\\ 
Blueprint & 40\% & 50\% & 50\% & 60\% & 50\% & 60\%\\ 
Landscape & 0\% & 30\% & 0\% & 30\% & 0\% & 40\%\\ 
\bottomrule
\end{tabularx}}
\end{table}

\begin{table}
\caption[Failure rate by document type for SuperPoint+SuperGlue]{Failure rate by document type for SuperPoint+SuperGlue.}
\label{tab:failure_rate_sg}
\centering
{\begin{tabularx}{0.49\textwidth}{X||c|c|c|c|c|c}
\toprule
Document & \multicolumn{2}{c|}{Medium} & \multicolumn{2}{c|}{Strong}& \multicolumn{2}{c}{Very Strong} \\
\cmidrule{2-7}
& Rotation & Scale & Rotation & Scale & Rotation & Scale \\
\midrule
Handwritten & 50\% & 90\% & 100\% & 100\%  & 100\% & 100\% \\ 
Typewritten & 0\% & 60\% & 20\% & 80\% & 10\% & 80\%\\ 
Blueprint & 0\% & 30\% & 20\% & 60\% & 0\% & 50\%\\ 
Landscape & 0\% & 20\% & 0\% & 30\% & 0\% & 30\%\\ 
\bottomrule
\end{tabularx}}
\end{table}

\subsection{Scale Robustness Experiment}

\begin{figure}
\centering
\begin{subfigure}{0.23\textwidth}
    \includegraphics[width=\textwidth]{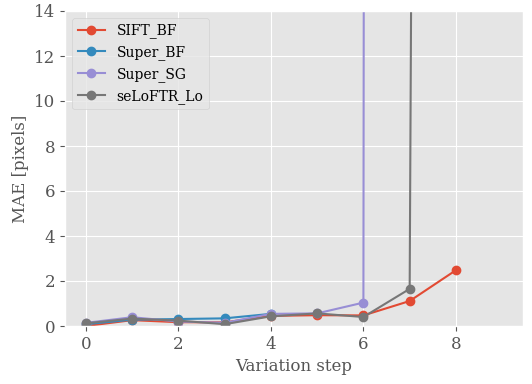}
    \caption{SIFT effectively reconstructs the snippet up to level 9 of downsize (90\% reduction), followed by SE2-LoFTR, which works up to level 8, and SP+SG, effective only until level 6.}
\end{subfigure}  
\hspace{0.01\textwidth}
\vspace{0.02\textwidth}
  \begin{subfigure}{0.23\textwidth}
    \includegraphics[width=\textwidth]{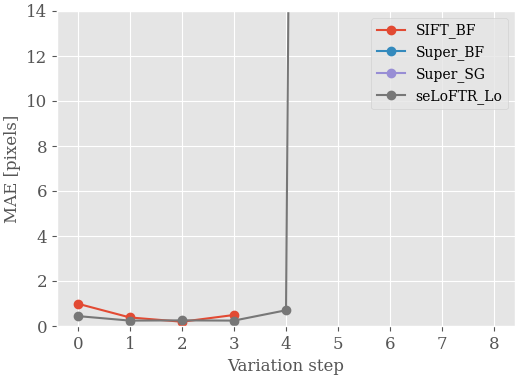}
    \caption{SE2-LoFTR manages to reconstruct the hardly damaged snippet up to level 4 of downsize, followed by SIFT, which stops at level 3; the other methods fail in every level.}
  \end{subfigure} 
  \caption{Scale Robustness Experiment: Handwritten document, MAE graphs for unfiltered images (left) and strong filtered images (see Fig.\ref{fig:filterlevel}) (right).}
    \label{fig:subfigures_scale}
  \end{figure}

Fragments, under four level of paper damage (see Fig. \ref{fig:filterlevel}), were downsized in ten increments, from their original size down to just 10\% of their original size. The landscape document emerged as the easiest to reconstruct across all levels of downsizing and filtering. This consistency underscores its relative simplicity compared to other document types. In contrast, documents with highly repetitive patterns presented the most significant challenges. Among these, the handwritten document proved to be the most difficult to reconstruct, followed by the typewritten documents in terms of difficulty. SE2-LoFTR effectively handled images with repetitive patterns across all filter levels, demonstrating its robustness. SIFT, while second-best for handwritten, performed slightly better for typewritten documents. SuperPoint+NN was effective only in the no-filter scenario; it failed to perform adequately in all the other damage levels, except with the landscape document. As the level of filter damage increased, SIFT’s scale invariance diminished, whereas SE2-LoFTR showed better adaptability in handling the more challenging hard and very hard damage categories (see Fig. \ref{fig:subfigures_scale}).

\subsection{Accuracy of the Automatic Positioning Approach}
\begin{figure}
\centering
\includegraphics[width=0.49\textwidth]{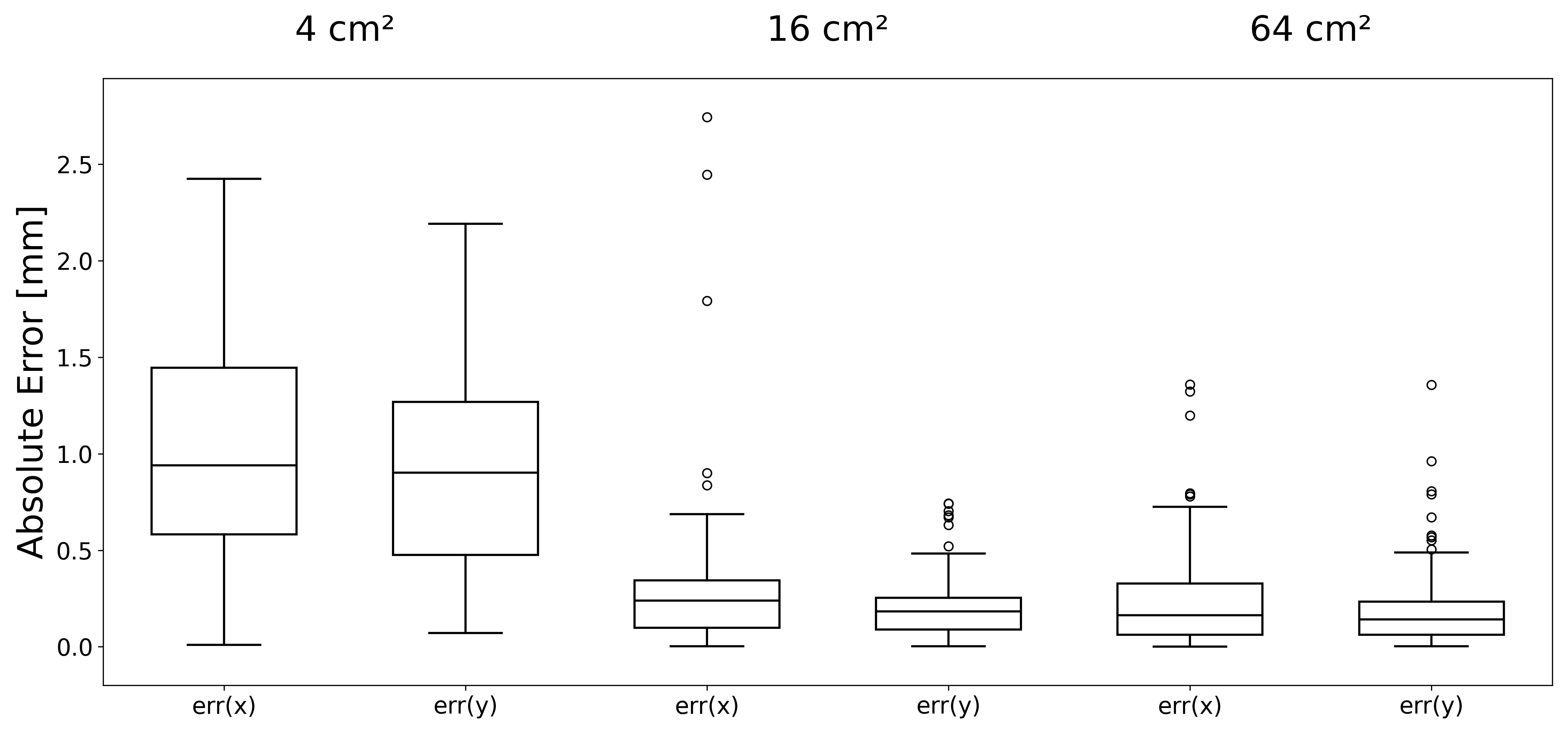}
\caption{Box plot illustrating the results of the experiment evaluating the accuracy of the UR3 cobot's automatic positioning of document fragments of varying sizes (4 cm², 16 cm², and 64 cm²). The plot presents the distribution of positioning errors for each fragment size divided into their X- and Y-components.}
\label{fig:exp_robot}
\end{figure}

In this experiment, we evaluate the repeatability of our automatic positioning approach using the UR3 cobot. To this end, three square fragments of differing sizes (4 cm², 16 cm², and 64 cm²) of 90g/m² paper with a robustly detectable feature were utilized. The fragments were each moved 100 times by the cobot from a defined constant start position to a defined target position. The height and speed of the movement were kept constant for all trials. To measure the accuracy of the positioning, the position of the feature was determined after each placement. The positioning errors are analysed by computing the mean coordinates of the detected features and determining the deviations from these mean values for each detected feature. The Root Mean Square Error (RMSE) is then calculated to quantify the overall accuracy. For the smallest fragment (4 cm²), the RMSE value is the highest (1.5254 mm); for the 16 cm² fragment the RMSE is 0.5735 mm and for the 64 cm² the RMSE is the lowest (0.4771 mm) (see Fig. \ref{fig:exp_robot}).

\section{Conclusion}

The primary objective of the development was to test and analyse the possibility of automating physical reconstruction tasks, especially of fragmented documents, using a cobot and machine vision methods.

As a result of this investigation, a vision system has been implemented that uses machine learning methods to automatically isolate paper fragments and determine the target position of the fragments in a dedicated target area. Experimentally, SE2-LoFTR was chosen as the reconstruction method as it provides robust matching results in the presence of high levels of damage (see \ref{tab:failure_rate_se2}). Existing limitations in rotational invariance can currently be circumvented by users when positioning in the workspace. The assumption that document types with font content are a comparatively difficult reconstruction task has also been confirmed. The developments also include a projector-based visual interface that can be used to visualise the target position for manual reconstruction. Based on a measurement and coordinate transformation, it is possible to use a cobot for the reconstruction task. This prototype already allows the automation of time-consuming tasks and would be capable of reconstructing large numbers of fragments, leaving the quality check to the experienced human eye.

\begin{figure}
\centering
\includegraphics[width=0.48\textwidth]{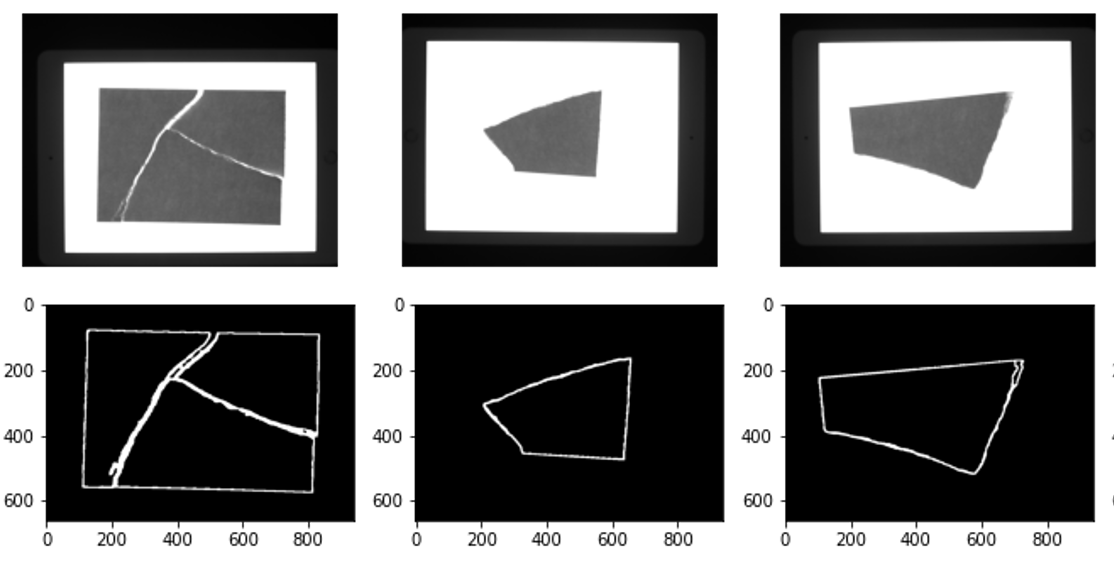}
\caption{Segmentation of fragments for the detection of inner and outer contours of the breaking edges. The contours were further preprocessed to fit the requirements of the cobot's path planning.}
\label{fig:contours}
\end{figure}

\section{Outlook and future work}

\begin{figure}
\centering
\includegraphics[width=0.45\textwidth]{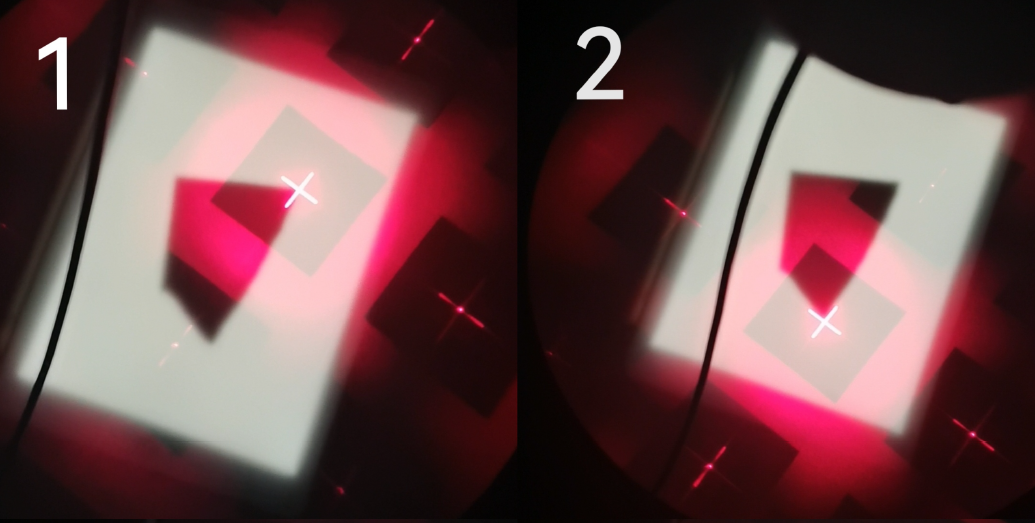}
\caption{Photo taken during the experiments showing the laser pointer, mounted on the Tool Center Point (TCP) of the UR3 cobot, as it follows the contours of a fragment. The cross marks the current position of the laser on the edge of the fragment.}
\label{fig:laser}
\end{figure}

It has been shown that some of the steps in the reconstruction work can be robustly automated by using cobots and machine vision techniques. Looking to the future, however, it would be possible to enable some further steps, such as the actual bonding or welding of the fragments, on the way to a fully automated system. In this context, some preliminary tests have already been carried out, in which a laser pointer was guided along the breaking points of the already positioned fragments with the help of the cobot used (see Fig. \ref{fig:laser}). The robot's path planning was based on automatic extraction of the inner and outer contours (see Fig. \ref{fig:contours}) using an image processing pipeline and transformed into the cobot's system using the calibration described. A classical interpolation method \cite{1240992} was used to derive the actual path. To validate the accuracy, the actual area of the fragment was compared to the area created by the laser path. The area-error was in average $1\%$. There was also an average deviation at significant points of $2.7$ mm. A value that leaves room for improvement.

\subsubsection{Acknowledgments}
We would like to thank the student project RoboGlue from the Institute of Automation Technology group of TU Berlin for the design of the suction system.

%
%
%
\bibliographystyle{splncs04}
\bibliography{literature}
%






\end{document}